\documentclass[11pt]{article}

\usepackage{arabtex}
\usepackage{utf8}
\usepackage{coling2018}
\usepackage{times}
\usepackage{url}
\usepackage{graphicx} 
\usepackage{pgfplots}

\usetikzlibrary{shapes,backgrounds}
\pgfplotsset{compat=1.14}

\usepackage[stable]{footmisc}

\date{}

\setcode{utf8}
\begin{document}

\title{\textnormal{Developing a Fine-Grained Corpus for a Less-resourced Language:\\the case of Kurdish}\footnotemark{}}\makeatletter{\renewcommand*{\@makefnmark}{}\footnotetext{\textsuperscript{*}This paper was peer-reviewed and published by WiNLP19. It originally appeared in the Proceedings of the 55th Annual Meeting of the Association for Computational Linguistics, pages 106--109, Florence, Italy, July 28th (see \url{https://www.aclweb.org/anthology/W19-3634/}), but as WiNLP's publication policy assumes the presented articles as non-archival elements and removes the full-texts after a while, we decided to archive the full-text through arxive.}\makeatother}

\author{
\begin{tabular}[t]{c c}
{Roshna Omer Abdulrahman} & Hossein Hassani\\
\textnormal{University of Kurdistan Hewl\^er} & \textnormal{University of Kurdistan Hewl\^er}\\
\textnormal{Kurdistan Region - Iraq} & 
\textnormal{Kurdistan Region - Iraq}\\
{\tt roshna.abdulrahman@ukh.edu.krd} & {\tt hosseinh@ukh.edu.krd}
\end{tabular}\vspace{0.15cm}\\
{\textbf {Sina Ahmadi}} \\
Insight Centre for Data Analytics\\
National University of Ireland Galway\\
Galway - Ireland\\ 
{\tt sina.ahmadi@insight-centre.org}}

\maketitle

\vskip 1.25cm

\begin{abstract}
	Kurdish is a less-resourced language consisting of different dialects written in various scripts. Approximately 30 million people in different countries speak the language. The lack of corpora is one of the main obstacles in Kurdish language processing. In this paper, we present KTC--the Kurdish Textbooks Corpus, which is composed of 31 K-12 textbooks in Sorani dialect. The corpus is normalized and categorized into 12 educational subjects containing 693,800 tokens (110,297 types). Our resource is publicly available for non-commercial use under the \texttt{CC BY-NC-SA 4.0} license\footnote{\url{https://creativecommons.org/licenses/by-nc-sa/4.0/}}.
\end{abstract}

\section{Introduction}
\label{intro}
Kurdish is an Indo-European language mainly spoken in central and eastern Turkey, northern Iraq and Syria, and western Iran. It is a less-resourced language \cite{salavati2018building}, in other words, a language for which general-purpose grammars and raw internet-based corpora are the main existing resources. The language is spoken in five main dialects, namely, Kurmanji (aka Northern Kurdish), Sorani (aka Central Kurdish), Southern Kurdish, Zazaki and Gorani \cite{haig2014introduction}.

\par Creating lexical databases and text corpora are essential tasks in natural language processing (NLP) development. Text corpora are knowledge repositories which provide semantic descriptions of words. The Kurdish language lacks diverse corpora in both raw and annotated forms \cite{esmaili2013building,hassani2018blark}. According to the literature, there is no domain-specific corpus for Kurdish.
 
\par In this paper, we present KTC, a domain-specific corpus containing K-12 textbooks in Sorani. We consider a domain as a set of related concepts, and a domain-specific corpus as a collection of documents relevant to those concepts \cite{mason2004cormet}. Accordingly, we introduce KTC as a domain-specific corpus because it is based on the textbooks which have been written and compiled by a group of experts, appointed by the Ministry of Education (MoE) of the Kurdistan Region of Iraq, for educational purposes at the K-12 level. The textbooks are selected, written, compiled, and edited by experts in each subject and also by language editors based on a unified grammar and orthography. This corpus was initially collected as an accurate source for developing a Sorani Kurdish spellchecker for scientific writing. KTC contains a range of subjects, and its content is categorized according to those subjects. Given the accuracy of the text from scientific, grammatical, and orthographic points of view, we believe that it is also a fine-grained resource. The corpus will contribute to various NLP tasks in Kurdish, particularly in language modeling and grammatical error correction.

In the rest of this paper, Section \ref{relatedwork} reviews the related work, Section \ref{thecorpus} presents the corpus, Section \ref{challenges} addresses the challenges in the project and, Section \ref{conclusion} concludes the paper.

\section{Related work}
\label{relatedwork}

Although the initiative to create a corpus for Kurdish dates back to 1998 \cite{gautier1998building}, efforts in creating machine-readable corpora for Kurdish are recent. The first machine-readable corpus for Kurdish is the Leipzig Corpora Collection which is constructed using different sources on the Web \cite{biemann2007leipzig}. Later, Pewan \cite{esmaili2013building} and Bianet \cite{ataman2018bianet} were developed as general-purpose corpora based on news articles. Kurdish corpora are also constructed for specific tasks such as dialectology \cite{malmasi2016subdialectal,hassani2018blark}, machine transliteration \cite{ahmadi2019wergor}, and part-of-speech (POS) annotation \cite{walther2010developing,walther2010fast}. However, to the best of our knowledge, currently, there is no domain-specific corpus for Kurdish dialects.

\section{The Corpus}
\label{thecorpus}
KTC is composed of 31 educational textbooks published from 2011 to 2018 in various topics by the MoE. We received the material from the MoE partly in different versions of Microsoft Word and partly in Adobe InDesign formats. In the first step, we categorized each textbook based on the topics and chapters. As the original texts were not in Unicode, we converted the content to Unicode. This step was followed by a pre-processing stage where the texts were normalized by replacing zero-width-non-joiner (ZWNJ) \cite{esmaili2013building} and manually verifying the orthography based on the reference orthography of the Kurdistan Region of Iraq. In the normalization process, we did not remove punctuation and special characters so that the corpus can be easily adapted our current task and also to future tasks where the integrity of the text may be required. \\

\begin{table}[ht]
\centering
\scalebox{1}{
\begin{tabular}{|l|c|c|c|c|}
\hline
Module title & Course level & \#Chapters & \#Tokens & \#Sentences \\ \hline \hline
Economics      &     12       &    7     &   32,823     &    1,023    \\ \hline
Genocide     &     10       &    8     &   16,243     &    670     \\ \hline
Geography    &     10       &    10    &   27,999     &    884     \\ \hline
History      &     10,12    &    20    &   79,845     &    2,065    \\ \hline
Human Rights &     10       &    5     &   11,527     &    340     \\ \hline
Kurdish      &  7,8,9,10,12 &    86    &   153,334    &    6,348    \\ \hline
Kurdology    &    10,11 (i) &    6     &   34,282     &    931     \\ \hline
Philosophy   &     11       &    6     &   21,953     &    549     \\ \hline
Physics      &  1,2,3,4 (i) &    30    &   111,032    &    4,022    \\ \hline
Theology &1,4,5,6,7,8,9,10,11,12& 191  &   115,349    &    3,661    \\ \hline
Sociology    &    8,9       &    42    &   68,044     &    2,082    \\ \hline
Social Study &     10       &    6     &   21,369     &    578     \\ \hline \hline
Total        &   31    &    417  &    693,800    &    23,153     \\ \hline
\end{tabular}
}
\caption{Statistics of the corpus - In the Course Level column, (i) represents Institute\protect\footnotemark.}
\label{SKETC_summary}
\end{table}

\footnotetext{The students could choose to go to the Institutes instead of High Schools after the Secondary School. The Institutes focus on professional and technical education aiming at training technicians.} 

\par As an experiment, we present the top 15 most used tokens of the textbooks in KTC, which are illustrated in Figure~\ref{fig:commontokens}. We observe that the most frequent tokens such as ({\small{\<ئابوورى >}} (economics), {\small{\<بازەرگانى >}} (business)) in economics, (=,$\times$ and {\small{\<وزەى >}} (energy)) in physics, and ({\small{\<خوداى >}} (god),  {\small{\<گەورە >}} (great) and  {\small{\<واتە >}} (meaning)) in theology are  conjunctions, prepositions, pronouns or punctuation. These are not descriptive of any one subject, while each subject's top tokens are descriptive of its content. The plot in Figure~\ref{fig:commontokens} follows Zipf's law to some extent, wherein the frequency of a word is proportional to its rank \cite{powers1998applications}. Here, not only the words but also the punctuation and special characters are also considered tokens (see Section~\ref{intro}).
\par The corpus is available at \url{https://github.com/KurdishBLARK/KTC}.\footnote{The materials of KTC are copyrighted. For more information refer to the provided link.} 
\pgfplotsset{width=7cm,compat=1.9}
\begin{figure}[ht]
    \centering
    \begin{tikzpicture}
    \begin{axis}[
        xlabel={Rank},
        ylabel={Frequency},
        legend pos=north west,
        ymajorgrids=true,
        grid style=dashed,
        x post scale = 2.7, 
        y post scale = 0.6
    ]
    \addplot[color=black,solid,thick,mark=*, mark options={fill=white}] 
        coordinates {
        
        (1,22846)(2,21335)(3,21321)(4,19772)(5,19589)(6,10893)(7,10379)(8,8827)(9,7285)(10,4482)(11,3979)(12,3116)(13,2253)(14,2005)(15,1927)
        }; 

    \node [below] at (axis cs:  1,22846) {.[dot]};
    \node [above] at (axis cs:  2,21335) {\small{{\<لە > [in]}}};
    \node [below] at (axis cs:  3,21321) {\small{{\<و > [and]}}};
    \node [below] at (axis cs:  4,19772) {)};
    \node [right] at (axis cs:  5,19589) {(};
    \node [above] at (axis cs:  6,10893) {:};
    \node [above] at (axis cs:  7,10379) {\small{{\<بە > [with]}}};
    \node [below] at (axis cs:  8,8827) {\small{{\<کە > [that]}}};
    \node [above] at (axis cs:  9,7285) {\small{{\<بۆ > [to]}}};
    \node [below] at (axis cs:  10,4482) {\small{{\<ئەو > [he, she]}}};
    \node [above] at (axis cs:  11,3979) {\small{{\<ئەم > [this]}}};
    \node [above] at (axis cs:  12,3116) {،};
    \node [above] at (axis cs:  13,2253) {...};
    \node [above] at (axis cs:  14,2005) {\small{{\<هەر > [any]}}};
    \node [right] at (axis cs:  15,1927) {\small{{\<وەک > [as]}}};
\end{axis}
\end{tikzpicture}
    \caption{Common tokens among textbook subjects.}
    \label{fig:commontokens}
\end{figure}
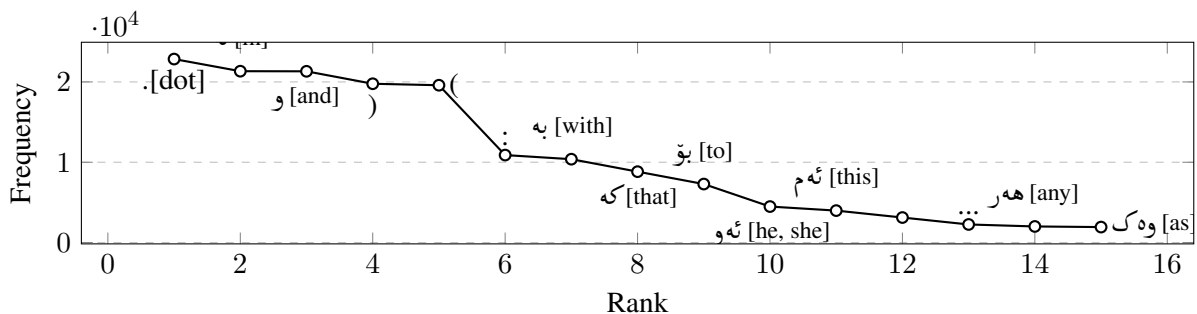




%













%
\section{Challenges}
\label{challenges}
Previously, researchers have addressed the challenges in Kurdish corpora development \cite{esmaili2013building,aliabadi2014towards,hassani2018blark}. We highlight two main challenges we faced during the KTC development. First, most of the written Kurdish resources have not been digitized \cite{hassani2019digital}, or they are either not publicly available or are not fully convertible. Second, Kurdish text processing suffers from different orthographic issues \cite{ahmadi2019wergor} mainly due to the lack of standard orthography and the usage of non-Unicode keyboards. Therefore, we carried out a semi-automatic conversion, which made the process costly in terms of time and human assistance.




\section{Conclusion}
\label{conclusion}
We presented KTC--the Kurdish Textbook Corpus, as the first domain-specific corpus for Sorani Kurdish. This corpus will pave the way for further developments in Kurdish language processing. We have mad the corpus available at \url{https://github.com/KurdishBLARK/KTC} for non-commercial use. We are currently working on a project on the Sorani spelling error detection and correction. As future work, we are aiming to develop a similar corpus for all Kurdish dialects, particularly Kurmanji.
\section*{Acknowledgments}
We would like to appreciate the generous assistance of the Ministry of Education of the Kurdistan Region of Iraq, particularly the General Directorate of Curriculum and Printing, for providing us with the data for the KTC corpus. Our special gratitude goes to Ms. Namam Jalal Rasheed and Mr. Kawa Omer Muhammad for their assistance in making the required data available and resolving of the copyright issues.
\bibliographystyle{acl}
\bibliography{winlp2019}

\end{document}